\documentclass[11pt,a4paper]{article}

\usepackage[T1]{fontenc}
\usepackage[utf8]{inputenc}
\usepackage{lmodern}
\usepackage{textcomp}
\usepackage[margin=2.5cm]{geometry}
\usepackage{microtype}
\usepackage{amsmath,amssymb}
\usepackage{graphicx}
\usepackage{xcolor}
\usepackage{booktabs}
\usepackage{multirow}
\usepackage{adjustbox}
\usepackage{rotating}
\usepackage[font=small,labelfont=bf]{caption}
\usepackage{authblk}
\usepackage[numbers,sort&compress]{natbib}
\usepackage{newunicodechar}
\usepackage[colorlinks=true,linkcolor=blue!60!black,citecolor=blue!60!black,urlcolor=blue!60!black]{hyperref}
\usepackage{comment}

\newunicodechar{±}{\ensuremath{\pm}}
\newunicodechar{×}{\ensuremath{\times}}
\newunicodechar{≤}{\ensuremath{\leq}}
\newunicodechar{≥}{\ensuremath{\geq}}
\newunicodechar{↑}{\ensuremath{\uparrow}}
\newunicodechar{↓}{\ensuremath{\downarrow}}
\newunicodechar{←}{\ensuremath{\leftarrow}}
\newunicodechar{∉}{\ensuremath{\notin}}
\newunicodechar{κ}{\ensuremath{\kappa}}
\newunicodechar{σ}{\ensuremath{\sigma}}
\newunicodechar{·}{\ensuremath{\cdot}}
\newunicodechar{°}{\ensuremath{^{\circ}}}

\title{\textbf{PCLSegNet: Anatomy-Aware Deep Learning Framework for MRI-based Pancreatic Cyst Segmentation and Size Assessment}}

\author[1,2]{Hanyi Zhang}
\author[1,3]{Tongyu Wang}
\author[1,4]{Manuel Debic}
\author[2]{Markus Wennmann}
\author[5]{Johannes Haubold}
\author[6]{Tamara Schikowski}
\author[7]{Wilfried Reichardt}
\author[7]{Fabian Bamberg}
\author[7]{Christopher L. Schlett}
\author[7]{Steffen Ringhof}
\author[8]{Thoralf Niendorf}
\author[8]{Tobias Pischon}
\author[9]{Michael Leitzmann}
\author[10]{Carsten Oliver Schmidt}
\author[10]{Mark Oliver Wielpütz}
\author[11]{Thomas Kroencke}
\author[11]{Josua A. Decker}
\author[12]{Stefan N. Willich}
\author[12,13,14]{Thomas Keil}
\author[15]{Rafael Mikolajczyk}
\author[1]{Miriam Klauß}
\author[1]{Hans-Ulrich Kauczor}
\author[1]{Philipp Mayer}
\author[1]{Bettina K. Budai\thanks{Corresponding author. Clinic for Diagnostic and Interventional Radiology, Heidelberg University Hospital, Im Neuenheimer Feld 130.3, 69120 Heidelberg, Germany. Email: \href{mailto:bettinakatalin.budai@med.uni-heidelberg.de}{bettinakatalin.budai@med.uni-heidelberg.de}}}

\affil[1]{Clinic for Diagnostic and Interventional Radiology, Heidelberg University Hospital, Medical Faculty of Heidelberg University, Heidelberg, Germany}
\affil[2]{Division of Radiology, German Cancer Research Center (DKFZ), Heidelberg, Germany}
\affil[3]{Department of Radiology, The Affiliated Hospital of Qingdao University, Qingdao, China}
\affil[4]{Division for Computational Radiology and Clinical AI, University Hospital Bonn, Faculty of Medicine at the University of Bonn, Bonn, Germany}
\affil[5]{Institute of Diagnostic and Interventional Radiology and Neuroradiology, University Hospital Essen, University Duisburg-Essen, Essen, Germany}
\affil[6]{IUF -- Leibniz Research Institute for Environmental Medicine, Düsseldorf, Germany}
\affil[7]{Department of Diagnostic and Interventional Radiology, Medical Center -- University of Freiburg, Faculty of Medicine -- University of Freiburg, Freiburg, Germany}
\affil[8]{Max Delbrück Center for Molecular Medicine in the Helmholtz Association, Berlin-Buch, Germany}
\affil[9]{Institute for Epidemiology and Preventive Medicine, University of Regensburg, Regensburg, Germany}
\affil[10]{Institute of Diagnostic Radiology and Neuroradiology, University Medicine Greifswald, Greifswald, Germany}
\affil[11]{Clinic for Diagnostic and Interventional Radiology, University Hospital Augsburg, Augsburg, Germany}
\affil[12]{Institute of Social Medicine, Epidemiology and Health Economics, Charité -- Universitätsmedizin Berlin, Berlin, Germany}
\affil[13]{Institute of Clinical Epidemiology and Biometry, University of Wuerzburg, Wuerzburg, Germany}
\affil[14]{State Institute of Health I, Bavarian Health and Food Safety Authority, Erlangen, Germany}
\affil[15]{Institute of Medical Epidemiology, Biometrics and Informatics, Medical Faculty of the Martin-Luther University Halle-Wittenberg, Halle (Saale), Germany}

\date{}

\begin{document}

\maketitle

\begin{abstract}
Early detection and follow-up of pancreatic cystic lesions (PCLs) are
crucial, as some of their subtypes, such as mucinous cystic neoplasms
(MCNs) and intraductal papillary mucinous neoplasms, carry a significant
risk of malignant transformation. Although most AI studies investigating
PCLs rely on computed tomography (CT) scans, magnetic resonance imaging
(MRI) offers superior soft tissue contrast, making it highly suitable
for PCLs assessment.

In this work, we propose an anatomy-aware two-stage deep learning
framework for automated PCLs segmentation on abdominal MRI scans. First,
a publicly available pancreas segmentation model is used to localize the
pancreas. Then, we introduce the anatomy-aware PCLSegNet, which
incorporates anatomical priors as guidance for pancreas and PCLs
segmentation. To address the class imbalance problem, we develop a
Class-Specific Error-Weighted (CSEW) loss function. Postprocessing is
applied through anatomical constraints. Our proposed method outperforms
multi-organ segmentation models such as TotalSegmentator, achieving
substantially higher Dice scores for pancreas segmentation. Moreover, it
also achieves a superior normalized surface Dice (NSD) and recall for
PCLs segmentation compared to other state-of-the-art models like
nnU-Net, while also notably reducing the false negative rate,
facilitating the automatic early detection and the noninvasive
imaging-based assessment of PCLs.
\end{abstract}


\section{Introduction}

Pancreatic cancer is the third most common cause of cancer-related
death, with incidence continuing to increase by approximately 1\%
annually over the past decade, and a contemporary 5-year survival rate
of only 13\%~\cite{siegel2026}. It can develop from pancreatic cystic
neoplasms~\cite{scholten2018}. While the incidental detection rate of
pancreatic cystic lesions (PCLs) on routine MRI is high (2.4--49.1\%),
clinical management remains challenging~\cite{ohtsuka2024}. Guidelines~\cite{ohtsuka2024,esgctp2018} recommend imaging surveillance or resection
based on size and morphology. Nevertheless, interobserver variability in
size measurement is significant (Cohen's kappa 0.59), and up to nearly
two-thirds of incidental PCLs are not included in routine reports~\cite{dunn2016,dippolito2018,megibow2017}. Automated tools are thus
essential to improve detection sensitivity, ensure reporting
consistency, and facilitate evaluation during follow-up.

Accurate PCL segmentation remains difficult due to the high anatomical
variability of the pancreas and the inherent heterogeneity of PCLs in
shape, size, and distribution. To date, only a limited body of
literature is available on PCL segmentation, and existing studies
exhibit several clinical constraints. In~\cite{wang2024}, 2D and 3D
U-Nets were trained on T2-weighted scans from 146 polycystic kidney
disease patients, achieving Dice similarity coefficients (DSC) of
0.70--0.80; however, their model's generalizability is limited by the
specific patient cohort. In~\cite{mazor2024}, a cascaded pipeline was
proposed for PCLs in coronal MRCP images, reaching a mean DSC of 0.80
for lesions \textgreater5 mm. Yet, the MRCP sequence is not routinely
included in abdominal MRI protocols, limiting its applicability for
incidental PCLs.

We hypothesized that an anatomy-aware two-stage deep learning framework
can improve PCL segmentation robustness by integrating pancreatic
contextual guidance and enable reliable automated size and location
monitoring. The purpose of this study was to develop and evaluate an
automated deep learning framework for the joint segmentation of the
pancreas and PCLs on T2-weighted MRI. Furthermore, we aimed to automate
the assessment of PCL size and anatomical location and to benchmark the
framework's performance against established state-of-the-art
segmentation models to demonstrate its potential clinical utility in the
assessment of incidental PCLs.

\section{Materials and Methods}

\subsection{Study Population and Dataset}

This retrospective study of prospectively acquired data was approved by
the local Institutional Review Board and the NAKO Use and Access
Committee (Application No. 204 \& 606). Written informed consent was
obtained from all participants prior to their inclusion in the German
National Cohort (NAKO)~\cite{bamberg2015}. The study was conducted in
accordance with the Declaration of Helsinki. While a number of previous
publications investigated the NAKO Study cohort, this is the first study
reporting the results of deep learning-based pancreatic cyst assessment.

This retrospective study included baseline abdominal T2W-MRI scans from
randomly selected 1,000 participants (Table~\ref{tab:demographics}), utilizing prospectively
acquired data from the NAKO study collected between January 2014 and
December 2019. Images were acquired at 3 Tesla (Skyra, Siemens
Healthineers, Erlangen, Germany) with an in-plane resolution of 1.4 ×
1.4 mm and a 5.0 mm slice thickness (6.0 mm voxel spacing, including a
1.0 mm interslice gap). To establish the reference standard ground
truth, all MRI scans were annotated and segmented by three experienced
radiologists: two with over 5 years and one with over 10 years of
experience in diagnostic radiology. The cohort was partitioned into a
training set and an independent test set using a 9:1 ratio. The training
set (900 participants aged 21-72 years; mean age of 54.5 ± 10.8 years;
49.7\% male) comprised 450 negative and 450 positive cases containing
636 PCLs. The test set (100 participants aged 22-72 years; mean age of
55.0 ± 11.2 years; 50\% male) included 50 negative and 50 positive cases
with 71 PCLs.

\subsection{First-Stage: Pancreas Segmentation}

\begin{figure}[t]
  \centering
  \includegraphics[width=0.85\textwidth]{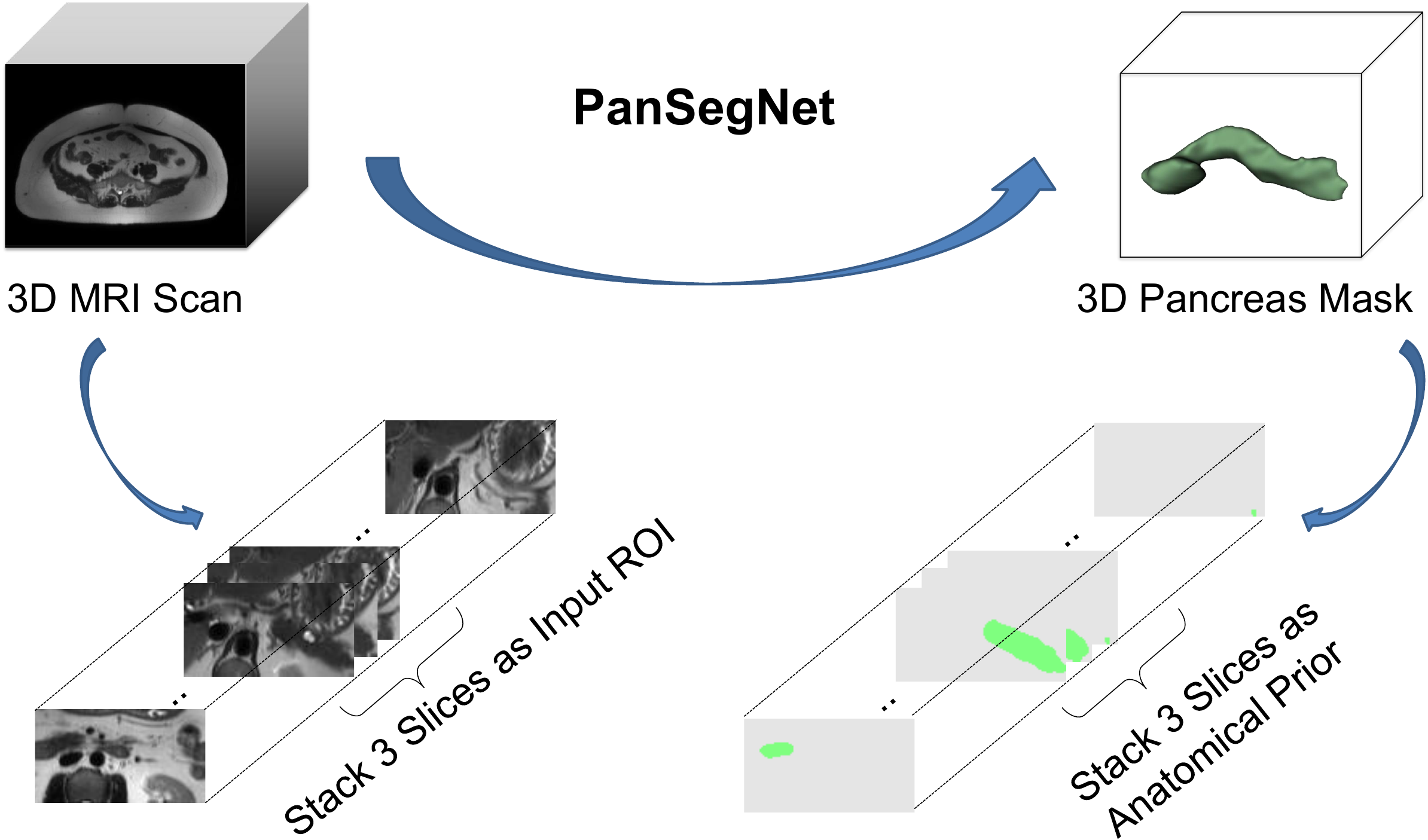}
  \caption{In stage 1, a 3D abdominal MRI scan is processed using PanSegNet to segment the pancreas, from which a bounding box is extracted and enlarged to form a region of interest (ROI). This ROI, along with the pancreas mask, is used to generate stacked 2.5D inputs.}
  \label{fig:stage1}
\end{figure}

To define the pancreatic region of interest (ROI), we employed a
pre-trained, pancreas-specific segmentation model, PanSegNet~\cite{zhang2025}, without additional fine-tuning. Predicted 3D masks
underwent morphological closing using an ellipsoidal kernel, followed by
the largest connected component selection. A 3D bounding box expanded by
3×3×2 voxels (x-, y-, z-axes) was extracted to preserve peripancreatic
anatomical context. For the subsequent stage, the cropped MRI volumes
and coarse pancreas masks were formatted into 2.5D inputs by stacking
three consecutive axial slices (Figure~\ref{fig:stage1}).

\subsection{Second-Stage: Anatomy-Aware PCL Segmentation}

\begin{figure}[t]
  \centering
  \includegraphics[width=0.85\textwidth]{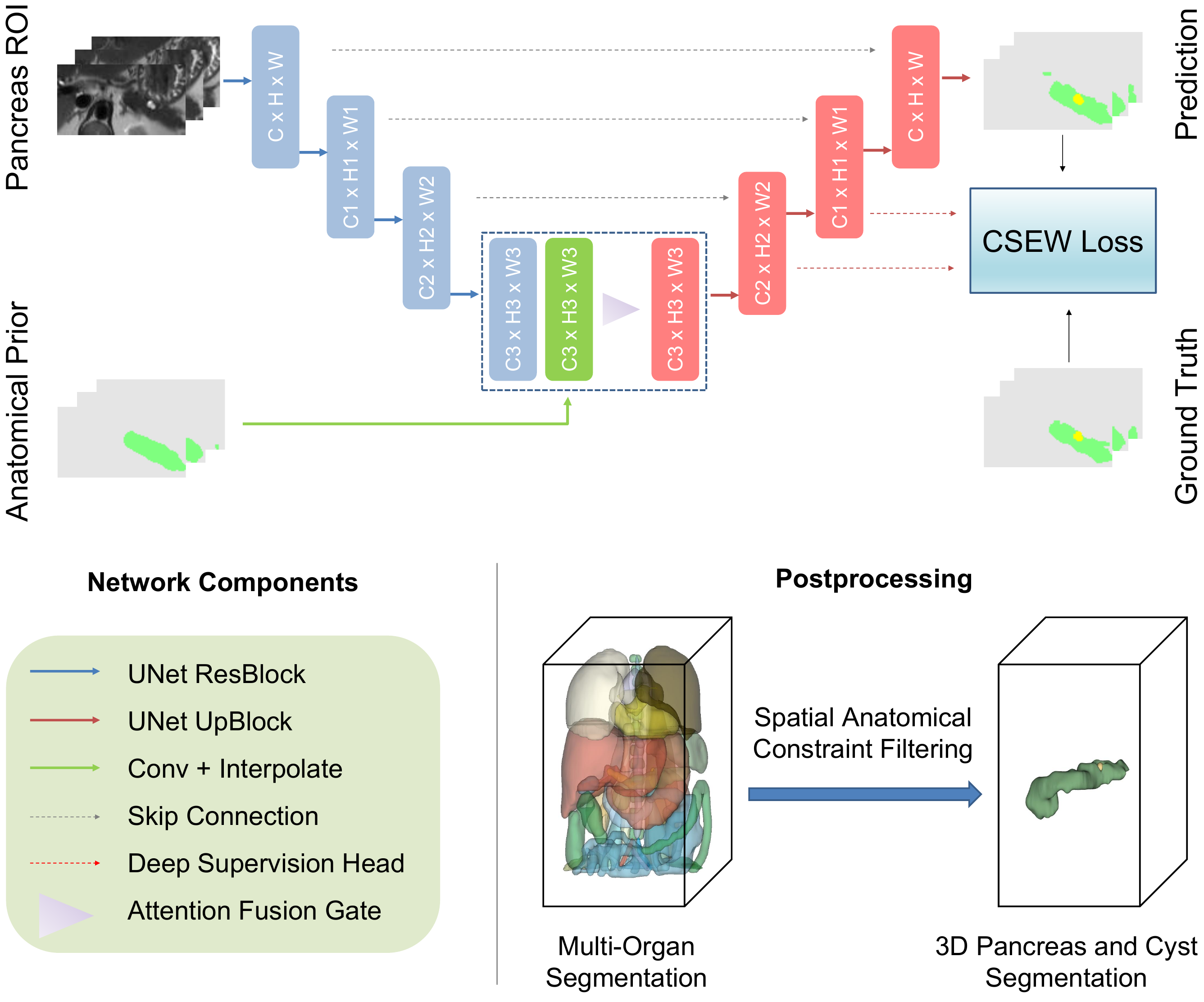}
  \caption{In stage 2, the pancreas ROI is processed by an image encoder, while the anatomical prior (predicted pancreas mask in stage 1) is passed through a convolutional block followed by interpolation. Both features are fused in the bottleneck using an Attention Fusion Gate and then decoded to generate the final segmentations. The model is supervised using the Class-Specific Error-Weighted loss. In the postprocessing step, predicted PCLs are refined using spatial anatomical constraints derived from multi-organ segmentation, which removes false positives located far from the pancreas or in adjacent organs.}
  \label{fig:stage2}
\end{figure}

A custom Anatomy-Aware 2.5D U-Net integrating the coarse pancreas mask
from the first stage as an anatomical prior was adopted for the second
stage to ensure computational efficiency while leveraging spatial
context (Figure~\ref{fig:stage2}). Prior to integration, the binary pancreas mask was
transformed into a continuous spatial feature map using a 3D Euclidean
Distance Transform, followed by Gaussian smoothing ($\sigma = 1.0$)
and {[}0, 1{]} intensity scaling. This pre-processed anatomical prior
was then concatenated with the image channels and processed through a
convolutional layer for channel adjustment. To align these anatomical
features with the deep image features, interpolation was applied to
match their spatial resolution at the bottleneck. These mask-derived
features were then fused with the image features via an Attention Fusion
Gate~\cite{oktay2018}, which guided the model to prioritize the
pancreatic parenchyma, suppressing extra-pancreatic noise.

Model training employed a five-times repeated five-fold cross-validation
strategy. We utilized the AdamW optimizer with an initial learning rate
of $1 \times 10^{- 2}$, following a polynomial decay scheduler
(power=0.9) over 1,000 epochs. To prevent overfitting, early stopping
was implemented based on validation performance. The framework was
developed using PyTorch Lightning and MONAI, executed on an NVIDIA
GeForce RTX 4080 SUPER GPU.

Data augmentations were applied during training to enhance
generalizability. Spatial augmentations, including random affine
transformations (rotation, scaling and translation), elastic deformation
and zooming, were applied to the input images, guidance masks and labels.
Intensity augmentations, including bias field, contrast adjustment and
Gaussian noise, were applied to the input images only.

Training was guided by deep supervision, integrating intermediate
outputs into the loss computation with a normalized weight ratio of
1:0.5:0.25. To address the class imbalance inherent in small lesion
segmentation, we introduced a Class-Specific Error-Weighted (CSEW) loss
for dynamically penalizing false-positive and false-negative predictions
for the PCL class by incorporating a pixel-wise weight map into both
Cross-Entropy and Dice loss components.

\subsection{Refined Postprocessing and Anatomical Constraints}

To clean the PCL mask predictions from artifacts, a multi-step
postprocessing strategy was implemented: (a) isolated single-voxel
regions were removed, (b) an anatomical filtering process was applied
using multi-organ masks generated by MRSegmentator~\cite{hntze2024},
(c) the pancreas mask was dilated to create an expanded search region,
and any predicted PCL component ≤ 10 voxels was discarded as artifacts
(e.g., lymph vessels) that had no overlap with the dilated mask.

\subsection{Automated PCL Assessment and Localization}

The maximum transverse diameter of the largest PCL was automatically
extracted from the final segmentation. For clinical localization, a
principal component analysis (PCA)-based geometric partitioning
algorithm was implemented by orienting the pancreas along its
longitudinal axis and dividing it into three equal segments as head,
body, and tail (detailed in Supplementary Section~\ref{app:pca}). This enabled
automated cyst localization and size-based assessment according to the
Kyoto guidelines~\cite{ohtsuka2024}.

\subsection{Model Evaluation and Statistical Analysis}

To identify true positive, false positive, and false negative PCLs, a
greedy matching algorithm was used to match the predicted PCL masks with
the ground truth annotations (detailed in Supplementary Section~\ref{app:matching}).
The performance of PCLSegNet was compared with several established
state-of-the-art segmentation architectures, including 2D nnU-Net, 3D
nnU-Net, Attention U-Net, UNETR, and V-Net. Segmentation performance was
evaluated at both the participant level and the PCL instance level using
the Dice similarity coefficient (DSC), normalized surface Dice (NSD),
and the 95th percentile Hausdorff distance (HD95). For PCL-level
evaluation, segmentation metrics were calculated for all annotated PCLs,
with false negative lesions assigned a score of zero. Detection
performance was assessed using precision, recall, and F1-score. At the
participant level, the True Positive Rate (TPR) was defined as the
proportion of positive participants correctly detected, whereas the True
Negative Rate (TNR) was defined as the proportion of negative
participants correctly identified. Continuous variables were expressed
as mean ± standard deviation. To assess the contribution of individual
components of the proposed framework, including the anatomy-aware
architecture, the CSEW loss, and the postprocessing strategy, an
ablation study was performed.

The agreement of maximum diameter measurements was assessed using Lin's
concordance correlation coefficient (CCC)~\cite{lin1989}. The classification of
PCLs into clinically relevant size categories was evaluated using
quadratic weighted $\kappa$~\cite{cohen1968} with 95\% confidence intervals (CIs). To
provide a human reference for model performance, an inter-observer
variability analysis was additionally conducted between two experienced
radiologists using the same diameter measurement and size-group
classification tasks. The accuracy of anatomical localization (e.g.,
pancreatic head versus non-head regions) was evaluated against expert
radiologist annotations.

\section{Results}

\subsection{Demographics}

Table~\ref{tab:demographics} summarizes the demographic and PCL characteristics of the
training and test datasets. The training dataset included 900
participants (50.3\% female; median age, 56 years {[}interquartile range
(IQR), 48--63{]}), while the test dataset consisted of 100 participants
(50.0\% female; median age, 56.5 years {[}IQR, 49--64{]}). The median
largest PCL size per participant was 7 mm in both datasets. The
distribution of PCL location was comparable between the training and
test datasets, with most lesions located in the pancreatic head or tail.
The average number of PCLs per positive participant was 1.41 in the
training dataset and 1.42 in the test dataset.

\begin{table}[t]
  \centering
  \caption{Demographic and PCL characteristics in training and test datasets. Median [interquartile range, 25th--75th percentile] values are reported for continuous variables.}
  \label{tab:demographics}
  \small
  \begin{tabular}{@{}lcc@{}}
    \toprule
    Characteristics & Training dataset ($n=900$) & Test dataset ($n=100$) \\
    \midrule
    Sex, n (\%) & & \\
    \quad Female & 453 (50.3) & 50 (50.0) \\
    \quad Male & 447 (49.7) & 50 (50.0) \\
    Age, years & 56 [48--63] & 56.5 [49--64] \\
    Largest cyst size per participant, mm & 7 [5--12] & 7 [5--10] \\
    Largest cyst location, n (\%) & & \\
    \quad Head & 183 (40.67) & 22 (44.00) \\
    \quad Body & 76 (16.89) & 6 (12.00) \\
    \quad Tail & 191 (42.44) & 22 (44.00) \\
    Average number of PCLs per positive participant & 1.41 & 1.42 \\
    \bottomrule
  \end{tabular}
\end{table}

\subsection{Evaluation of Pancreas Segmentation Performance}

To obtain preliminary pancreas masks for defining the target
field-of-view on the full 3D abdominal MRI scans, we compared the
PanSegNet with the two other publicly available 3D segmentation models:
TotalSegmentator MRI~\cite{dantonoli2025} and MRSegmentator~\cite{hntze2024} in a zero-shot manner, i.e., without fine-tuning. The
results uncovered varying generalization capabilities (Table~\ref{tab:pancreas}); the
PanSegNet achieved the highest zero-shot DSC of 0.91 on the test set,
representing a substantial improvement of 0.42 over TotalSegmentator MRI
(DSC=0.49) and 0.19 over MRSegmentator (DSC=0.72).

\begin{table}[t]
  \centering
  \caption{Performance comparison of different models for pancreas segmentation (mean DSC $\pm$ standard deviation). Models are grouped by usage: zero-shot versus in-house trained models. Model size is reported as the total number of parameters in millions (M). Best results are shown in bold and second-best results are underlined.}
  \label{tab:pancreas}
  \small
  \begin{tabular}{@{}llrll@{}}
    \toprule
    Model usage & Method & Param (M) & Input scope & DSC \\
    \midrule
    Zero-shot & TotalSegmentator MRI & 176.50 & Full 3D scan & 0.49 \\
     & MRSegmentator & 88.20 & Full 3D scan & 0.72 \\
     & PanSegNet & 43.90 & Full 3D scan & \textbf{0.91} \\
    \midrule
    In-house trained & 2D nnU-Net & 31.00 & 2D ROI & 0.89 \\
     & 3D nnU-Net & 41.20 & 3D ROI & \underline{0.90} \\
     & Attention U-Net & 23.40 & 3D ROI & 0.83 ± 0.07 \\
     & UNETR & 121.00 & 3D ROI & 0.80 ± 0.08 \\
     & V-Net & 45.60 & 3D ROI & 0.82 ± 0.07 \\
     & PCLSegNet (ours) & 8.20 & 2.5D ROI & 0.88 ± 0.00 \\
    \bottomrule
  \end{tabular}
\end{table}

\begin{figure}[t]
  \centering
  \includegraphics[width=\textwidth]{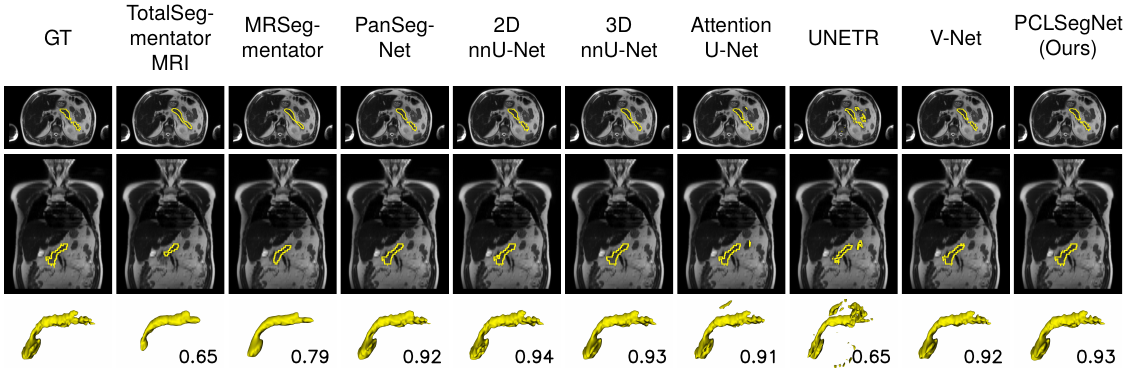}
  \caption{Qualitative comparison of pancreas segmentation results on an example test case. For visual demonstration of the pancreas segmentation performance, the segmentation results of the nine trained deep learning models are shown together with the ground truth mask annotated by the radiologist on a representative example case. While the basic shape and location of the pancreas is captured on all predicted pancreas masks in this example, the predictions of TotalSegmentator MRI and MRSegmentator constantly appeared smooth, allowing for only a coarse segmentation and losing the texture details. In comparison, both 2D and 3D nnU-Net, as well as VNet and our proposed PCLSegNet model, achieved superior results by better capturing surface granularity. Although Attention U-Net and UNETR achieved relatively good Dice scores, they noticeably predicted false positive regions, leading to less accurate results. Rows from top to bottom: axial slices, coronal slices, and 3D surface reconstructions with Dice similarity coefficients.}
  \label{fig:pancreas}
\end{figure}

Moreover, for the task of pancreas and PCL joint segmentation, the
performance of our proposed PCLSegNet was compared to five
state-of-the-art models trained on our dataset: 2D and 3D nnU-Net~\cite{isensee2021}, Attention U-Net~\cite{oktay2018}, UNETR~\cite{hatamizadeh2022}, and V-Net~\cite{milletari2016}. The segmentation masks
were predicted on the pancreas ROI-cropped MRIs. The results of the
models' performance are summarized in Table~\ref{tab:pancreas}, and a representative
example is visualized in Figure~\ref{fig:pancreas}. The consistently high Dice scores
demonstrate excellent overall pancreas segmentation performance across
these models. However, the predictions of TotalSegmentator MRI and
MRSegmentator were generally too smooth, losing surface granularity
details. In contrast, the Attention U-Net and UNETR produced noticeably
more predicted false positive areas.

\subsection{Evaluation of PCL Segmentation Performance}

The detailed results of the models' overall performance are summarized
in Table~\ref{tab:pcl}. During participant-level evaluation, PCLSegNet achieved the
highest TPR of 84.0\% (42/50), corresponding to a 6\% improvement over
the second-best model, 3D nnU-Net. At the PCL level, where false
negative lesions were assigned a score of zero to jointly assess
detection and segmentation performance, PCLSegNet consistently achieved
the best overall segmentation performance. Specifically, it yielded the
highest DSC (0.52 ± 0.02), NSD (0.68 ± 0.03), recall (0.77 ± 0.03), and
inverse HD95 (0.23 ± 0.00), indicating both superior lesion detection
sensitivity and more accurate boundary delineation.

\begin{sidewaystable}
  \centering
  \caption{Performance comparison of different models for PCL segmentation and detection at the participant and PCL levels. For PCL-level metrics, segmentation metrics were calculated for all annotated PCLs, with false negative lesions assigned a score of zero. Results are presented as mean $\pm$ standard deviation where applicable. Participant-level metrics include true positive rate (TPR), true negative rate (TNR), and overall accuracy. PCL-level metrics include Dice similarity coefficient (DSC), normalized surface Dice (NSD), inverse 95th percentile Hausdorff distance (HD95), precision, recall, and F1-score. Arrows indicate the direction of superior performance: ↑ (higher is better). Best results are shown in bold and second-best results are underlined.}
  \label{tab:pcl}
  \footnotesize
  \setlength{\tabcolsep}{4pt}
  \begin{tabular}{@{}lccccccccccc@{}}
    \toprule
     & \multicolumn{3}{c}{Participant-level metrics} & \multicolumn{7}{c}{PCL-level metrics} \\
    \cmidrule(lr){2-4}\cmidrule(l){5-11}
    Method & TPR ↑ & TNR ↑ & Overall Acc ↑ & DSC ↑ & NSD ↑ & Inv.\ HD95 ↑ & Precision ↑ & Recall ↑ & F1-score ↑ & Avg.\ TP ↑ \\
    \midrule
    2D nnU-Net & 0.72 & \textbf{0.96} & \underline{0.84} & 0.43 & 0.54 & \underline{0.20} & 0.83 & 0.63 & 0.72 & 45.0 \\
    3D nnU-Net & \underline{0.78} & \underline{0.94} & \textbf{0.86} & \underline{0.50} & \underline{0.62} & \textbf{0.23} & \textbf{0.87} & \underline{0.68} & \textbf{0.76} & \underline{48.0} \\
    Attention U-Net & 0.66 ± 0.01 & 0.92 ± 0.03 & 0.79 ± 0.02 & 0.34 ± 0.01 & 0.43 ± 0.01 & 0.15 ± 0.01 & 0.81 ± 0.06 & 0.54 ± 0.01 & 0.64 ± 0.03 & 38.0 \\
    UNETR & 0.71 ± 0.04 & 0.16 ± 0.06 & 0.44 ± 0.01 & 0.31 ± 0.04 & 0.43 ± 0.06 & 0.16 ± 0.02 & 0.26 ± 0.03 & 0.66 ± 0.05 & 0.37 ± 0.02 & 46.8 \\
    V-Net & 0.52 ± 0.16 & 0.91 ± 0.06 & 0.72 ± 0.05 & 0.26 ± 0.09 & 0.35 ± 0.12 & 0.11 ± 0.04 & \underline{0.84 ± 0.11} & 0.44 ± 0.14 & 0.55 ± 0.09 & 31.0 \\
    PCLSegNet (ours) & \textbf{0.84 ± 0.02} & 0.74 ± 0.07 & 0.79 ± 0.03 & \textbf{0.52 ± 0.02} & \textbf{0.68 ± 0.03} & \textbf{0.23 ± 0.00} & 0.74 ± 0.06 & \textbf{0.77 ± 0.03} & \underline{0.75 ± 0.02} & \textbf{54.4} \\
    \bottomrule
  \end{tabular}
\end{sidewaystable}

While the overall segmentation metrics demonstrate the superior
performance of PCLSegNet, lesion size is a clinically important factor
that substantially influences segmentation difficulty. Therefore, we
further evaluated model performance across the four Kyoto guideline size
categories (Table~\ref{tab:size}). Overall, PCLSegNet demonstrated the most
consistent and robust performance across all lesion size groups.
Notably, the proposed framework achieved the highest or jointly highest
recall in every size category, ranging from 0.52 for lesions ≤5 mm to
1.00 for lesions \textgreater20 mm. Similarly, PCLSegNet consistently
delivered the strongest or second-strongest segmentation performance,
achieving NSD values of 0.47, 0.71, 0.73, and 0.90 across the four size
groups, respectively. Importantly, the model maintained superior
performance even for the smallest lesions (≤5 mm), which are generally
the most challenging to detect and segment. The performance advantage
became increasingly pronounced for larger lesions (\textgreater20 mm),
where PCLSegNet achieved an NSD of 0.90 ± 0.01 and an inverse HD95 of
0.24 ± 0.02, substantially outperforming all competing methods. These
findings indicate that the proposed framework maintains accurate
boundary delineation across the full spectrum of clinically relevant PCL
sizes.

\begin{table}[t]
  \centering
  \caption{PCL-level segmentation performance across different PCL size groups (Kyoto classification). Results are presented as mean $\pm$ standard deviation where applicable. Best results are shown in bold and second-best results are underlined.}
  \label{tab:size}
  \footnotesize
  \setlength{\tabcolsep}{4pt}
  \begin{tabular}{@{}lcccccc@{}}
    \toprule
     & \multicolumn{6}{c}{PCL-level metrics} \\
    \cmidrule(l){2-7}
    Method & Dice ↑ & NSD ↑ & Inv.\ HD95 ↑ & Recall ↑ & TP (\#) ↑ & FN (\#) ↓ \\
    \midrule
    \multicolumn{7}{@{}l}{\textit{Size group ≤ 5 mm} (total PCLs: 13)} \\
    \quad 2D nnU-Net & 0.25 & 0.29 & \underline{0.12} & 0.31 & 4 & 9 \\
    \quad 3D nnU-Net & 0.14 & 0.20 & 0.09 & 0.23 & 3 & 10 \\
    \quad Attention U-Net & 0.09 ± 0.02 & 0.13 ± 0.01 & 0.06 ± 0.01 & 0.15 ± 0.00 & 2.00 ± 0.00 & 11.00 ± 0.00 \\
    \quad UNETR & \underline{0.26 ± 0.05} & \underline{0.41 ± 0.09} & \textbf{0.19 ± 0.03} & \underline{0.51 ± 0.08} & \underline{6.60 ± 1.02} & \underline{6.40 ± 1.02} \\
    \quad V-Net & 0.09 ± 0.08 & 0.13 ± 0.12 & 0.06 ± 0.05 & 0.14 ± 0.12 & 1.80 ± 1.60 & 11.20 ± 1.60 \\
    \quad PCLSegNet (ours) & \textbf{0.32 ± 0.02} & \textbf{0.47 ± 0.03} & \textbf{0.19 ± 0.04} & \textbf{0.52 ± 0.03} & \textbf{6.80 ± 0.40} & \textbf{6.20 ± 0.40} \\
    \midrule
    \multicolumn{7}{@{}l}{\textit{Size group > 5 mm to ≤ 10 mm} (total PCLs: 25)} \\
    \quad 2D nnU-Net & 0.40 & 0.52 & 0.17 & 0.64 & 16 & 9 \\
    \quad 3D nnU-Net & \textbf{0.53} & \underline{0.67} & \underline{0.26} & \underline{0.72} & \underline{18} & \underline{7} \\
    \quad Attention U-Net & 0.39 ± 0.02 & 0.51 ± 0.03 & 0.19 ± 0.02 & 0.57 ± 0.03 & 14.20 ± 0.75 & 10.80 ± 0.75 \\
    \quad UNETR & 0.44 ± 0.04 & 0.59 ± 0.05 & 0.23 ± 0.02 & 0.71 ± 0.05 & 17.80 ± 1.33 & 7.20 ± 1.33 \\
    \quad V-Net & 0.25 ± 0.11 & 0.33 ± 0.15 & 0.12 ± 0.06 & 0.38 ± 0.16 & 9.60 ± 3.93 & 15.40 ± 3.93 \\
    \quad PCLSegNet (ours) & \underline{0.51 ± 0.03} & \textbf{0.71 ± 0.04} & \textbf{0.27 ± 0.01} & \textbf{0.82 ± 0.05} & \textbf{20.60 ± 1.36} & \textbf{4.40 ± 1.36} \\
    \midrule
    \multicolumn{7}{@{}l}{\textit{Size group > 10 mm to ≤ 20 mm} (total PCLs: 30)} \\
    \quad 2D nnU-Net & 0.53 & 0.66 & \underline{0.23} & \underline{0.73} & \underline{22} & \underline{8} \\
    \quad 3D nnU-Net & \textbf{0.63} & \textbf{0.75} & \textbf{0.28} & \textbf{0.80} & \textbf{24} & \textbf{6} \\
    \quad Attention U-Net & 0.41 ± 0.02 & 0.51 ± 0.02 & 0.17 ± 0.01 & 0.65 ± 0.03 & 19.40 ± 1.02 & 10.60 ± 1.02 \\
    \quad UNETR & 0.25 ± 0.05 & 0.35 ± 0.06 & 0.11 ± 0.02 & 0.65 ± 0.06 & 19.40 ± 1.85 & 10.60 ± 1.85 \\
    \quad V-Net & 0.35 ± 0.08 & 0.46 ± 0.11 & 0.13 ± 0.03 & 0.57 ± 0.15 & 17.00 ± 4.38 & 13.00 ± 4.38 \\
    \quad PCLSegNet (ours) & \underline{0.58 ± 0.03} & \underline{0.73 ± 0.04} & 0.21 ± 0.01 & \textbf{0.80 ± 0.04} & \textbf{24.00 ± 1.10} & \textbf{6.00 ± 1.10} \\
    \midrule
    \multicolumn{7}{@{}l}{\textit{Size group > 20 mm} (total PCLs: 3)} \\
    \quad 2D nnU-Net & 0.48 & 0.59 & 0.07 & \textbf{1.00} & \textbf{3} & \textbf{0} \\
    \quad 3D nnU-Net & \underline{0.59} & \underline{0.63} & \underline{0.10} & \textbf{1.00} & \textbf{3} & \textbf{0} \\
    \quad Attention U-Net & 0.22 ± 0.04 & 0.28 ± 0.06 & 0.03 ± 0.01 & 0.80 ± 0.16 & 2.40 ± 0.49 & 0.60 ± 0.49 \\
    \quad UNETR & 0.06 ± 0.02 & 0.10 ± 0.02 & 0.04 ± 0.00 & \textbf{1.00 ± 0.00} & \textbf{3.00 ± 0.00} & \textbf{0.00 ± 0.00} \\
    \quad V-Net & 0.22 ± 0.13 & 0.29 ± 0.14 & 0.04 ± 0.02 & \underline{0.87 ± 0.27} & \underline{2.60 ± 0.80} & \underline{0.40 ± 0.80} \\
    \quad PCLSegNet (ours) & \textbf{0.80 ± 0.00} & \textbf{0.90 ± 0.01} & \textbf{0.24 ± 0.02} & \textbf{1.00 ± 0.00} & \textbf{3.00 ± 0.00} & \textbf{0.00 ± 0.00} \\
    \bottomrule
  \end{tabular}
\end{table}

\begin{figure}[p]
  \centering
  \includegraphics[width=0.78\textwidth]{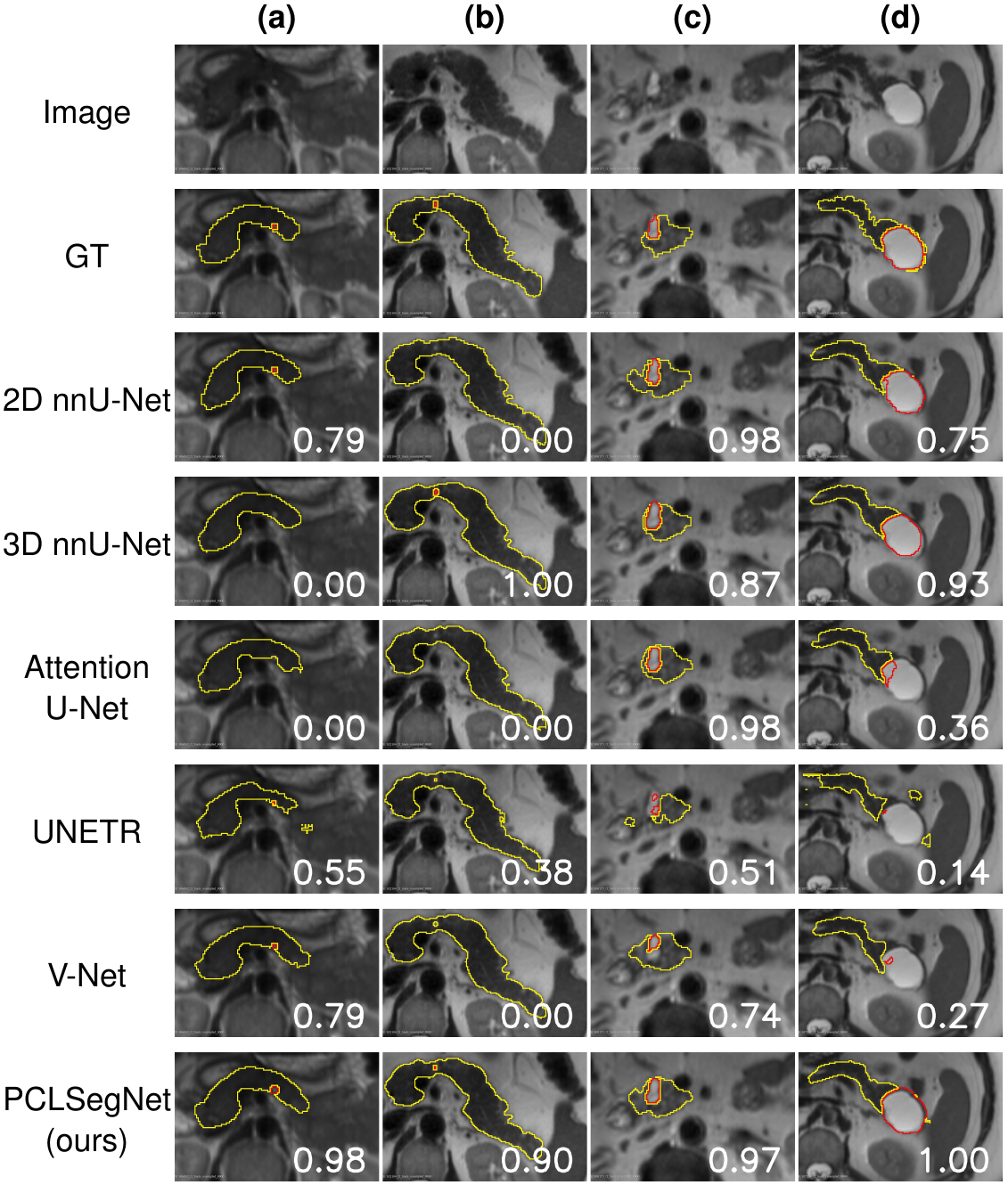}
  \caption{Visual comparison of segmentation results. Rows show the largest cross-section of the original MRI image, the ground truth (GT) segmentation, and the predictions of the five state-of-the-art models versus our proposed PCLSegNet. Columns (a)--(d) represent four PCLs from diameter groups with different clinical relevance (classified according to the Kyoto guidelines into ≤ 5 mm vs.\ 5--10 mm vs.\ 10--20 mm vs.\ > 20 mm). The normalized surface Dice (NSD) score is displayed in the bottom-right corner. Please note that the images show the largest cross-sectional axial slice, but the metrics are calculated from 3D segmentations.
  Column (a) represents a PCL from the size-group ≤ 5 mm located in the pancreatic body. Notably, both the 3D nnU-Net and Attention U-Net failed to segment this PCL, misclassifying it as pancreas parenchyma. While the 2D nnU-Net and V-Net achieved an NSD of 0.79, PCLSegNet achieved 0.98, demonstrating near-perfect segmentation.
  A similar trend is evident in column (b) (size-group > 5 mm to ≤ 10 mm), where PCLSegNet maintains strong performance with an NSD of 0.90, whereas the 2D nnU-Net, Attention U-Net, and V-Net all failed to detect or segment the target PCL.
  For larger PCLs, as shown in columns (c) and (d), all models successfully detected the lesions, but with varying degrees of segmentation accuracy. In column (c), the PCL was measured 16 mm large by the radiologist. The largest diameter measured by PCLSegNet was 17 mm, while Attention U-Net and 2D nnU-Net measured 16 mm and 17 mm, respectively, showing close agreement with the ground truth. Conversely, 3D nnU-Net and V-Net yielded less accurate measurements of 19 mm and 12 mm, respectively, despite their relatively high NSD scores. In example (d), featuring an exceptionally large PCL (maximum diameter of 44 mm), PCLSegNet achieved a perfect segmentation with an NSD of 1.00, demonstrating superior volumetric capture compared to all baseline models.}
  \label{fig:pcl}
\end{figure}

To provide a visual interpretation of these quantitative findings,
representative examples from each size category are shown in Figure~\ref{fig:pcl},
illustrating the superior segmentation accuracy of PCLSegNet across both
small and large lesions.

\subsection{Evaluation of PCL Detection Performance}

Beyond superior boundary delineation, our model also showed superior
detection performance compared to state-of-the-art methods.
Specifically, PCLSegNet achieved a remarkably higher recall of 77\%
(54.4/71), outperforming the second-best model (3D nnU-Net), which
achieved a recall of 68\% (48.0/71), by 9.0\%. In comparison, UNETR and
2D nnU-Net ranked third and fourth in this metric, respectively. Their
relatively lower recall indicates reduced sensitivity in detecting PCLs,
which may limit their utility in clinical scenarios where minimizing
false negatives is highly desirable. In contrast, PCLSegNet achieved the
second-highest F1-score (within 1\% of the top-performing model),
demonstrating its ability to maintain a robust balance between precision
and recall.

To better understand the nature of the remaining detection errors, a
radiologist (with over 5 years of experience) performed a confidence
assessment of all FP and FN detections from one seed prediction. Each
case was categorized as either low- or high-confidence based on the
review of the axial T2-weighted HASTE sequence alone. Interestingly,
90.9\% of FN cases were rated as low confidence, suggesting that most
missed lesions represent borderline findings whose presence is
challenging to determine even for experienced readers using only the
available sequence. This observation indicates that a substantial
proportion of the model's FN detections may correspond to inherently
ambiguous cases rather than obvious missed cysts. For FP detections, the
radiologist reported high confidence in 84.1\% of cases, identifying
them as non-cystic anatomical structures such as segments of the bile
duct, the duodenum, or other peripancreatic tissues. This finding
suggests that the majority of FP errors arise from specific anatomical
confounders rather than random false alarms. Addressing these
anatomically challenging false-positive detections remains a limitation
of the current framework and will be an important direction for future
work.

These results demonstrate that PCLSegNet can detect a greater number of
PCLs while simultaneously providing highly accurate segmentation masks.
Moreover, the observed detection errors are largely clinically
interpretable and reveal clear avenues for further model refinement
(Table~\ref{tab:pcl}).

\subsection{Ablation Study}

The contribution of the individual components of the proposed framework
was evaluated through an ablation study (Table~\ref{tab:ablation}). Compared with the
baseline U-Net trained using DiceCE loss, the anatomy-aware PCLSegNet
combined with the CSEW loss substantially improved segmentation
performance, increasing the DSC from 0.39 to 0.52, the NSD from 0.52 to
0.69, and the recall from 0.63 to 0.78. These improvements indicate
enhanced lesion detection sensitivity and more accurate boundary
delineation. However, the increased sensitivity was accompanied by a
reduction in precision (0.80 vs. 0.67), suggesting a higher number of
false positive detections. To address this limitation, the proposed
postprocessing strategy was applied, resulting in a marked improvement
in precision from 0.67 to 0.74 while maintaining a high recall of 0.77.
Consequently, the complete PCLSegNet framework achieved the highest
overall F1-score of 0.75, demonstrating the complementary benefits of
the anatomy-aware design, CSEW loss, and postprocessing strategy.

\begin{table}[t]
  \centering
  \caption{Ablation study. Starting from a baseline U-Net trained with DiceCE, the effects of the anatomy-aware architecture, CSEW, and postprocessing strategy were evaluated. Results are presented as mean $\pm$ standard deviation. Best results are shown in bold and second-best results are underlined.}
  \label{tab:ablation}
  \footnotesize
  \setlength{\tabcolsep}{3.5pt}
  \begin{tabular}{@{}lllcccccc@{}}
    \toprule
    Model & Loss & Postproc. & DSC & NSD & Inv.\ HD95 & Precision & Recall & F1 \\
    \midrule
    U-Net & DiceCE & No & \underline{0.39 ± 0.01} & 0.52 ± 0.02 & 0.18 ± 0.01 & \textbf{0.80 ± 0.04} & 0.63 ± 0.02 & 0.70 ± 0.02 \\
    PCLSegNet & CSEW & No & \textbf{0.52 ± 0.02} & \textbf{0.69 ± 0.03} & \textbf{0.24 ± 0.00} & 0.67 ± 0.07 & \textbf{0.78 ± 0.03} & \underline{0.72 ± 0.03} \\
    PCLSegNet & CSEW & Yes & \textbf{0.52 ± 0.02} & \underline{0.68 ± 0.03} & \underline{0.23 ± 0.00} & \underline{0.74 ± 0.06} & \underline{0.77 ± 0.03} & \textbf{0.75 ± 0.02} \\
    \bottomrule
  \end{tabular}
\end{table}

\subsection{Evaluation of Automated PCL Assessment and Anatomical Localization}

To assess the clinical reliability of automated size measurements, the
agreement between predicted and ground truth sizes was tested using two
complementary metrics. First, CCC was employed to measure the fidelity
of maximum diameter. Second, the quadratic weighted κ was used to
evaluate the classification of PCLs into clinically relevant size
categories. As summarized in Table~\ref{tab:agreement}, PCLSegNet demonstrated superior
performance in maximum diameter estimation, achieving a CCC of 0.97
(95\% CI: 0.81, 0.99). This indicates nearly-perfect agreement and
surpasses both 2D nnU-Net (0.92) and 3D nnU-Net (0.93).

\begin{table}[t]
  \centering
  \caption{Agreement analysis of diameter measurement and size-based classification: Lin's concordance correlation coefficient (CCC) and quadratic weighted $\kappa$ with their respective 95\% confidence intervals [95\% CI]. Best results are shown in bold and second-best results are underlined.}
  \label{tab:agreement}
  \small
  \begin{tabular}{@{}lcc@{}}
    \toprule
     & Maximum diameter & Size-based grouping$^{*}$ \\
    Model & CCC [95\% CI] & Weighted $\kappa$ [95\% CI] \\
    \midrule
    2D nnU-Net & 0.92 [0.58, 0.97] & 0.64 [0.43, 0.81] \\
    3D nnU-Net & \underline{0.93} [0.56, 0.97] & \underline{0.67} [0.43, 0.82] \\
    Attention U-Net & 0.72 [0.42, 0.79] & 0.48 [0.18, 0.71] \\
    UNETR & 0.15 [$-$0.01, 0.43] & 0.28 [0.07, 0.49] \\
    V-Net & 0.64 [0.38, 0.73] & 0.48 [0.15, 0.72] \\
    PCLSegNet (ours) & \textbf{0.97} [0.81, 0.99] & \textbf{0.71} [0.57, 0.81] \\
    \bottomrule
  \end{tabular}

  \smallskip
  \begin{minipage}{0.8\textwidth}\footnotesize
  CCC: Lin's concordance correlation coefficient; CI: confidence interval.\\
  $^{*}$Size-based grouping as ≤ 5 mm vs.\ > 5 mm to ≤ 10 mm vs.\ > 10 mm to ≤ 20 mm vs.\ > 20 mm according to the Kyoto guidelines.
  \end{minipage}
\end{table}

Regarding size classification, PCLSegNet achieved a substantial κ of
0.71 (95\% CI: 0.57, 0.81), outperforming 2D nnU-Net (0.64) and 3D
nnU-Net (0.67). These results confirm that PCLSegNet provides highly
reliable size monitoring and robust categorical assignment, essential
for standardized clinical risk assessment.

To provide a human reference for model performance, an inter-observer
variability analysis was additionally conducted between two experienced
radiologists. The inter-observer agreement for maximum diameter
measurements achieved a CCC of 0.91 (95\% CI: 0.53, 0.96), indicating
substantial agreement. For Kyoto guideline size-group classification,
the inter-observer agreement reached a quadratic weighted κ of 0.55
(95\% CI: 0.35, 0.70), corresponding to moderate agreement. Compared
with the observer agreement, PCLSegNet achieved a higher CCC of 0.97 and
a higher weighted κ of 0.71, demonstrating excellent reproducibility for
both diameter measurement and clinically relevant size-group
classification.

Beyond lesion segmentation, the clinical utility of our framework was
further evaluated through automated anatomical localization of the
largest PCL of each participant, specifically distinguishing between the
pancreatic head and non-head (body and tail) regions. On the independent
test set, the proposed pipeline achieved an overall localization
accuracy of 80.0\%. Notably, the model demonstrated high sensitivity for
lesions in the pancreatic head, correctly localizing 86.4\% (19/22) of
these cases. For non-head lesions, the model achieved a sensitivity of
75.0\% (21/28) and a precision of 95.5\% (21/22). The high localization
fidelity, particularly for head-located PCLs, underscores the potential
of our framework to assist in standardized clinical reporting and risk
stratification according to guidelines.

\subsection{Model Size}

Finally, it is important to highlight that our proposed model is
markedly more lightweight than established state-of-the-art segmentation
architectures. PCLSegNet utilizes only 8.2 M parameters, which is nearly
one-third the size of Attention U-Net and one-fifth that of 3D nnU-Net,
as shown in Table~\ref{tab:pancreas}. Consequently, PCLSegNet is highly suitable for
deployment in clinical environments with limited computational
resources.

\section{Discussion}

In this study, we addressed the critical clinical challenge of
interobserver variability and high radiologist non-reporting rates in
evaluating incidental PCLs on T2W MRI~\cite{dunn2016,dippolito2018,megibow2017}. To mitigate these limitations, we
evaluated a fully automated, anatomy-aware deep learning framework
tailored for robust PCL segmentation, precise size assessment, and
anatomical localization. Our primary findings demonstrated that our 2.5D
network architecture, which explicitly integrates an anatomy-aware prior
via an attention fusion gate, successfully bridges the gap between
spatial contextual information and computational efficiency. By
capturing essential multi-slice contextual guidance while markedly
reducing model complexity, this specialized pipeline achieves technical
segmentation accuracy comparable to heavy state-of-the-art models
alongside significantly higher clinical detection sensitivity. The
ablation study further confirmed that the anatomy-aware design and
postprocessing strategy both contributed substantially to the observed
performance gains. Crucially, the automated diameter measurements and
categorical risk assignments yielded substantial agreement with the
expert radiological reference standard, underscoring the pipeline's
potential to standardize opportunistic screening in routine clinical
workflows.

The inclusion of the NSD metric in our evaluation highlights the
clinical nuance of our boundary assessment. Because PCLs typically
occupy a substantially smaller volume than the surrounding pancreatic
parenchyma, traditional voxel-based metrics like the DSC can be overly
sensitive to minor boundary deviations, masking true clinical utility.
By incorporating a tolerance threshold that accounts for surface
boundary alignment, our boundary evaluations confirmed that the proposed
framework delivers high geometric fidelity along the lesion-tissue
interface. This accurate delineation may facilitate volumetric tracking
and longitudinal assessment of morphologic changes associated with
increased malignancy risk.

Current diagnostic and clinical management guidelines heavily rely on
size-based risk stratification to determine whether a patient requires
intensive imaging surveillance or surgical resection. The exceptionally
high CCC achieved by our model for maximum diameter estimation indicates
substantial linear fidelity with expert measurements. Similarly, the
robust weighted kappa value underscores that our automated size-based
grouping aligns closely with the categorical consensus of experienced
abdominal radiologists. This capability addresses a major pitfall in
routine reporting, where geometric variance between different readers
often leads to inconsistent follow-up recommendations. Importantly, the
agreement achieved by PCLSegNet exceeded the observed inter-observer
agreement for both diameter measurement and size-group classification,
suggesting that part of the remaining discrepancy may reflect inherent
measurement variability rather than model limitations.

Beyond pure dimensional metrics, automated anatomical localization
(specifically head versus non-head partitioning) provides crucial
spatial context. The high sensitivity observed for lesions within the
pancreatic head is clinically significant. Lesions in the pancreatic
head may have greater clinical relevance because of their association
with biliary obstruction and pancreatitis.

Our current study has several limitations. First, our model was
developed and validated exclusively using the T2-weighted HASTE
sequence. It has not been tested for the other MRI techniques, such as
fat-suppressed T2W and T1W sequences, which radiologists typically use
in conjunction for comprehensive diagnosis. Second, while the model
successfully performs PCL detection, segmentation, localization, size
measurement, and size categorization, it currently does not support
histopathological subtyping. Future investigations should focus on (1)
incorporating the segmentation of surrounding critical anatomical
structures, such as the bile duct and the main pancreatic duct, to
provide richer contextual information for diagnosis, (2) validating
robustness on external clinical datasets across diverse T2-weighted
sequences, and (3) extending the pipeline to support longitudinal
follow-up monitoring of PCL dynamics by evaluating size-changes and
morphological signs of malignant transformation.

In conclusion, our anatomy-aware two-stage deep learning framework
provides an efficient and highly reliable automated solution for
MRI-based pancreatic cyst evaluation. PCLSegNet matches the geometric
segmentation performance of heavy state-of-the-art architectures while
delivering superior detection sensitivity through a markedly lightweight
design. By accurately automating maximum diameter measurements,
categorical size stratification, and anatomical localization, this
framework demonstrates substantial agreement with expert radiologists,
offering a valuable tool to enhance screening efficiency and reporting
consistency for incidental pancreatic lesions.

\section*{Acknowledgements}
This project was conducted with data (Application No. 204 \& 606)
from the German National Cohort (NAKO) (www.nako.de). The NAKO is funded
by the Federal Ministry of Education and Research (BMBF) {[}project
funding reference numbers: 01ER1301A/B/C, 01ER1511D, 01ER1801A/B/C/D and
01ER2301A/B/C{]}, federal states of Germany and the Helmholtz
Association, the participating universities and the institutes of the
Leibniz Association. We thank all participants who took part in the NAKO
study and the staff of this research initiative.

The authors acknowledge support by the state of Baden-Württemberg
through bwHPC and the German Research Foundation (DFG) through grant
INST 35/1597-1 FUGG.

\section*{Ethics Statement}
This study was approved by the Ethics Committee of the Medical Faculty
Heidelberg, Heidelberg University (approval number S-256/2021; approved
April 12, 2021). The study was conducted in accordance with the
Declaration of Helsinki. All NAKO participants provided written informed
consent for participation and secondary use of their imaging and health
data for approved research projects.
Participant confidentiality and privacy were protected in accordance
with GDPR and applicable German data protection regulations.

\section*{Data Availability Statement}
The datasets analyzed during the current study are not publicly
available due to ethical, legal, and privacy restrictions associated
with human participant data. Access to the data may be considered upon
reasonable request and subject to approval by the relevant data access
committees, institutional policies, and applicable data protection
regulations.

\bibliographystyle{unsrtnat}
\bibliography{refs}

\clearpage
\appendix
\renewcommand{\thesection}{S\arabic{section}}
\renewcommand{\thetable}{S\arabic{table}}
\setcounter{table}{0}
\section*{Supplementary Material}
\addcontentsline{toc}{section}{Supplementary Material}

\section{Automated Geometric Localization of PCLs Using Principal Component Analysis}
\label{app:pca}

To establish a reproducible reference for categorizing pancreatic cystic lesions (PCLs) into ``head'' and ``non-head'' regions, we implemented an automated 3D geometric localization algorithm based on Principal Component Analysis (PCA). The workflow consists of four steps.

First, the 3D spatial coordinates $(x, y, z)$ of all foreground voxels were extracted from the segmented binary pancreas mask. PCA was then performed on these coordinates to determine the major longitudinal axis of the pancreas, defined by the first principal component (PC1). This axis provides a standardized centerline extending from the pancreatic head through the body to the tail.

Next, the 3D centroid of each segmented PCL was calculated and projected onto the patient-specific PC1 axis. Finally, the extent of the pancreas along PC1 was divided into three segments of equal length, corresponding to the head, body, and tail. A PCL was classified as a ``head lesion'' if its projected centroid fell within the head segment; otherwise (i.e., within the body or tail segment), it was classified as a ``non-head lesion.''

\section{Greedy Matching for Multi-Instance Assignment}
\label{app:matching}

To evaluate PCL detection performance, a greedy matching process is performed between predicted and ground truth PCL instances. The step-by-step logic of this optimization routine is formally outlined in Algorithm~S1. An instance is considered a True Positive (TP) if it has a non-zero overlap with a ground truth instance and has not been previously matched. Unmatched ground truth instances are counted as False Negatives (FN), while unmatched predicted instances are False Positives (FP). Precision, Recall, and the F1-score are reported using the following standard formulations:
\begin{equation*}
  \text{Precision} = \frac{TP}{TP + FP}, \qquad
  \text{Recall} = \frac{TP}{TP + FN}, \qquad
  \text{F1} = \frac{2 \cdot \text{Precision} \cdot \text{Recall}}{\text{Precision} + \text{Recall}}.
\end{equation*}

\begin{table}[h]
  \centering
  \small
  \begin{tabular}{@{}rl@{}}
    \multicolumn{2}{@{}l}{\textbf{Algorithm S1:} Greedy matching for PCL instance detection.} \\
    \toprule
    1: & Initialize $\mathcal{M}_{gt} \leftarrow \emptyset$ and $\mathcal{M}_{pred} \leftarrow \emptyset$ \\
    2: & \textbf{for} each ground truth PCL $c_{gt}$ in GT \textbf{do} \\
    3: & \quad Find the predicted PCL $c_{pred}$ with the largest overlap with $c_{gt}$ \\
    4: & \quad \textbf{if} $\mathrm{overlap}(c_{gt}, c_{pred}) > 0$ \textbf{and} $c_{pred} \notin \mathcal{M}_{pred}$ \textbf{then} \\
    5: & \qquad Mark $c_{gt}$ and $c_{pred}$ as matched \\
    6: & \qquad Add $c_{gt}$ to $\mathcal{M}_{gt}$ \\
    7: & \qquad Add $c_{pred}$ to $\mathcal{M}_{pred}$ \\
    8: & \quad \textbf{end if} \\
    9: & \textbf{end for} \\
    10: & TP $\leftarrow$ number of matched pairs \\
    11: & FN $\leftarrow$ number of unmatched GT PCLs \\
    12: & FP $\leftarrow$ number of unmatched predicted PCLs \\
    \bottomrule
  \end{tabular}
\end{table}

\end{document}